\documentclass[letter, 10pt, conference]{ieeeconf}
\IEEEoverridecommandlockouts

\usepackage{cite}

\usepackage{amsmath,amssymb,amsfonts}
\usepackage{algorithmic}
\usepackage{graphicx}
\usepackage{subcaption}
\usepackage{pdfpages}
\usepackage{textcomp}
\usepackage{xcolor}
\usepackage{bm}
\usepackage{booktabs}
\usepackage{svg}
\svgsetup{inkscapelatex=false}
\usepackage{tikz}
\usepackage{hyperref}
\def\BibTeX{{\rm B\kern-.05em{\sc i\kern-.025em b}\kern-.08em
    T\kern-.1667em\lower.7ex\hbox{E}\kern-.125emX}}
\usetikzlibrary{arrows,decorations.pathmorphing,backgrounds,fit,positioning,calc,shapes}

\usepackage[linesnumbered,ruled,algo2e,resetcount]{algorithm2e}
\SetAlFnt{\small}
\SetAlCapFnt{\small}
\SetAlCapNameFnt{\small}

\begin{document}
\bstctlcite{MyBSTcontrol}

\title{\LARGE \bf AZRA: Extending the Affective Capabilities of \\ Zoomorphic Robots using Augmented Reality
\thanks{This work is supported by the University of Glasgow’s EPSRC Impact Acceleration Account (Grant Number EP/X525716/1).}
}

\author{\authorblockN{
Shaun Macdonald$^{a}$,
Salma ElSayed$^{b}$,
Mark McGill$^{a}$
}
\thanks{$^{a}$University of Glasgow, Scotland}
\thanks{$^{b}$Abertay University, Scotland}
}
\maketitle

\begin{figure}[h!]
  \centering
  \includegraphics[width=1.0\linewidth]{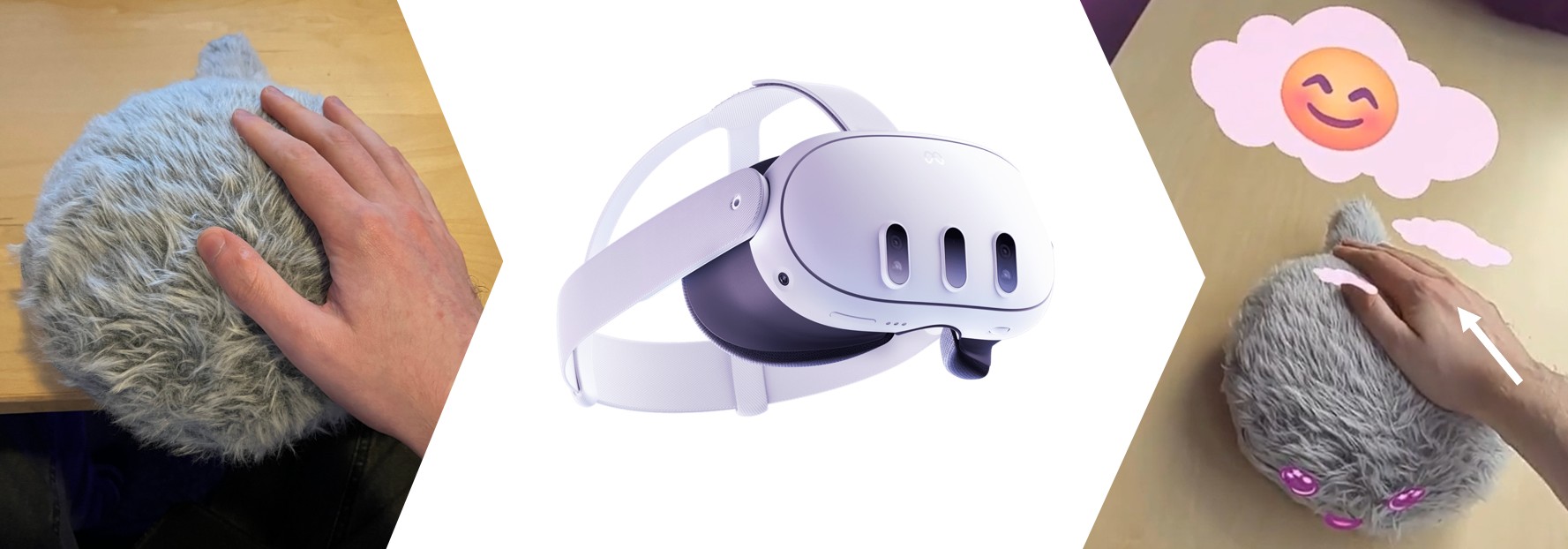}
  \caption{The AZRA framework uses pass-through AR to augment an existing zoomorphic robot with new affective displays, interactions, and a slowly evolving CME that models agency and appraises interactions.}
    \label{fig:teaser}
\end{figure}

\begin{abstract}
Zoomorphic robots could serve as accessible and practical alternatives for users unable or unwilling to keep pets.
However, their affective interactions are often simplistic and short-lived, limiting their potential for domestic adoption.
In order to facilitate more dynamic and nuanced affective interactions and relationships between users and zoomorphic robots we present AZRA, a novel augmented reality (AR) framework that extends the affective capabilities of these robots without physical modifications.
To demonstrate AZRA, we augment a zoomorphic robot, Petit Qoobo, with novel emotional displays (face, light, sound, thought bubbles) and interaction modalities (voice, touch, proximity, gaze).
Additionally, AZRA features a computational model of emotion to calculate the robot's emotional responses, daily moods, evolving personality and needs. 
We highlight how AZRA can be used for rapid participatory prototyping and enhancing existing robots, then discuss implications on future zoomorphic robot development.

\end{abstract}

\section{Introduction}

Zoomorphic robots have the potential to emulate the benefits of animal interactions, provide comfort~\cite{borgstedt_soothing_2024}, and facilitate affective touch and social interaction in contexts such as medical and care settings~\cite{Wada2007}. 
Zoomorphic robots offer practical advantages for everyday users who face financial, accessibility, or logistical challenges in owning pets.
However, their domestic use is hampered by limited affective interactions and intelligence that impede their ability to establish engaging and enduring pet-like relationships, instead placing a burden of personification upon users~\cite{Young2011a, Katsuno2022, Sefidgar2016}. 
\par
Zoomorphic robots have seen comparatively little advancement over the past two decades.
The majority of research exploring the application of relatively simple robots in healthcare settings \cite{Hudson2020, Jeong2018} and there have been calls to move beyond simple short-term affective interactions in HRI~\cite{Tanevska2020, cavallo_emotion_2018}.
Robots like Paro\footnote{Paro Robot http://www.parorobots.com/}, Qoobo\footnote{Petit Qoobo https://qoobo.info/index-en} and Joy for All companions\footnote{Joy for All Cat: joyforall.com/products/companion-cats} feature a small set of simple and predictable affective interactions, while other robots such as Moflin\footnote{Moflin: https://www.moflin.com MofLife App: tinyurl.com/moflinapp} rely on external apps to explicitly communicate affective information.
Furthermore, creating and modifying robots is time-consuming and resource-intensive, impeding research of their affective design space to simple prototypes~\cite{Loffler2018, macdonald_evaluating_2024}.
\par
Seeking to address these challenges, we present the Augmenting Zoomorphic Robotics with Affect (AZRA) framework.
The core concept of AZRA is that new affective expressions, sensing, interactions and emotion modelling are projected onto a robot, mediated by the user's AR head-mounted display (HMD).
Meanwhile, the robot itself remains unchanged, allowing new affective capabilities to be implemented regardless of the original robot's capabilities, without requiring users to replace their robot.
This approach has the potential to reduce e-waste, facilitate rapid prototyping, and looks toward a future of everyday AR and its potential impact on human-robot interaction~\cite{Suzuki2022, macdonald_evaluating_2024}. 

\par
AZRA allows robots to use additional holistic multimodal displays that communicate a greater variety of emotional and practical meanings. 
It further extends the robot's ability to sense affective user interactions and proxemic information, enhancing the potential for varied meaningful interactions.
Finally, AZRA incorporates an adapted computational model of emotion (CME) \cite{dias_fatima_2014} that mediates the robot's short-term emotional responses, slowly evolving personality, daily moods, and internal needs.
These additions are designed to facilitate more fulfilling, engaging and dynamic affective relationships between user and robot, potentially making them more suitable as domestic alternatives.
\par
The primary contribution of this report is a novel AR framework to augment zoomorphic robots with new emotional interaction and intelligence.
We describe its core functionality and the new displays, interactions and behaviours we have implemented, as well as our motivations.
We conclude by outlining our next steps to evaluate and utilise the framework and highlight wider implications on the future affective design of zoomorphic robots and beyond.

\section{AR for Affective Zoomorphic Interaction} \label{sec:methods}

This section details our apparatus, approach, and three ways that AZRA extends a robot's affective capabilities.

\subsection{Apparatus and Approach} 
\begin{figure}[h!]
  \centering
  \includegraphics[width=0.38\linewidth]{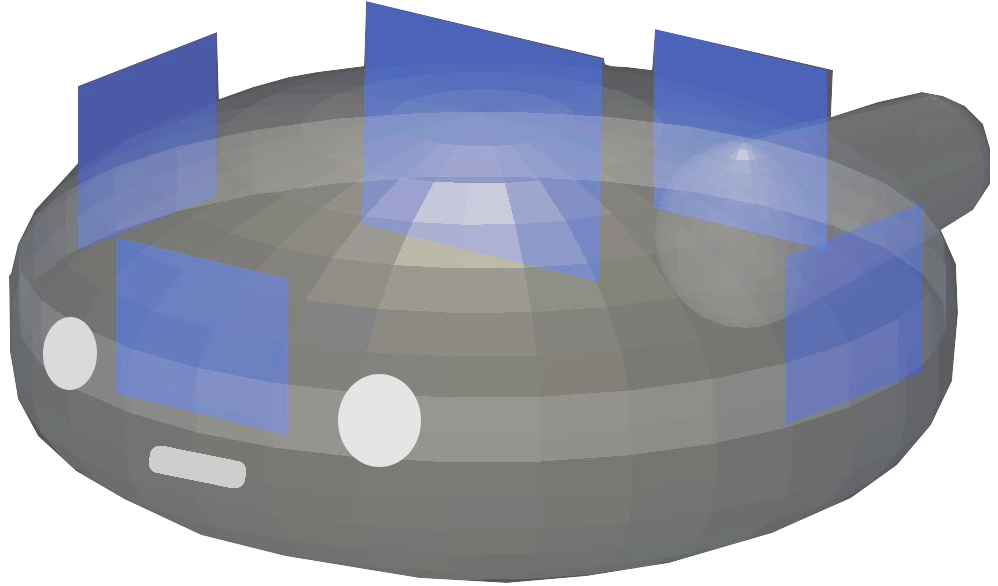}
  \caption{A transparent virtual copy is overlaid on top of the robot using AR, acting as a surface and locus for new affective displays and interaction. Blue surfaces indicate colliders used to detect hand-tracked touch.}
    \label{fig:virtualcopy}
\end{figure}

We chose the robot Petit Qoobo to demonstrate the AZRA framework. 
It features a soft vibrotactile heartbeat and an expressive moving tail. 
We selected this robot due to a simple abstract design that provided ample opportunities for varied augmentations.
To augment the robot, we created a transparent 3D model of the robot using Blender\footnote{Blender: https://www.blender.org/} and overlaid this onto the robot in pass-through AR (see Fig. \ref{fig:virtualcopy}).  
This virtual copy then acts as a locus and surface for new affective displays and user interactions.
Currently, the virtual copy is manually placed using AR hand tracking, though future iterations could automate this process via object detection or physical trackers.
The HMD used was the Meta Quest 3, while the framework was built using Unity 2022.3.47f1 and runs in real-time via Meta Quest Link.

\subsection{Augmenting the Robot's Affective Display Modalities}
\begin{figure}[h!]
  \centering
  \includegraphics[width=0.94\linewidth]{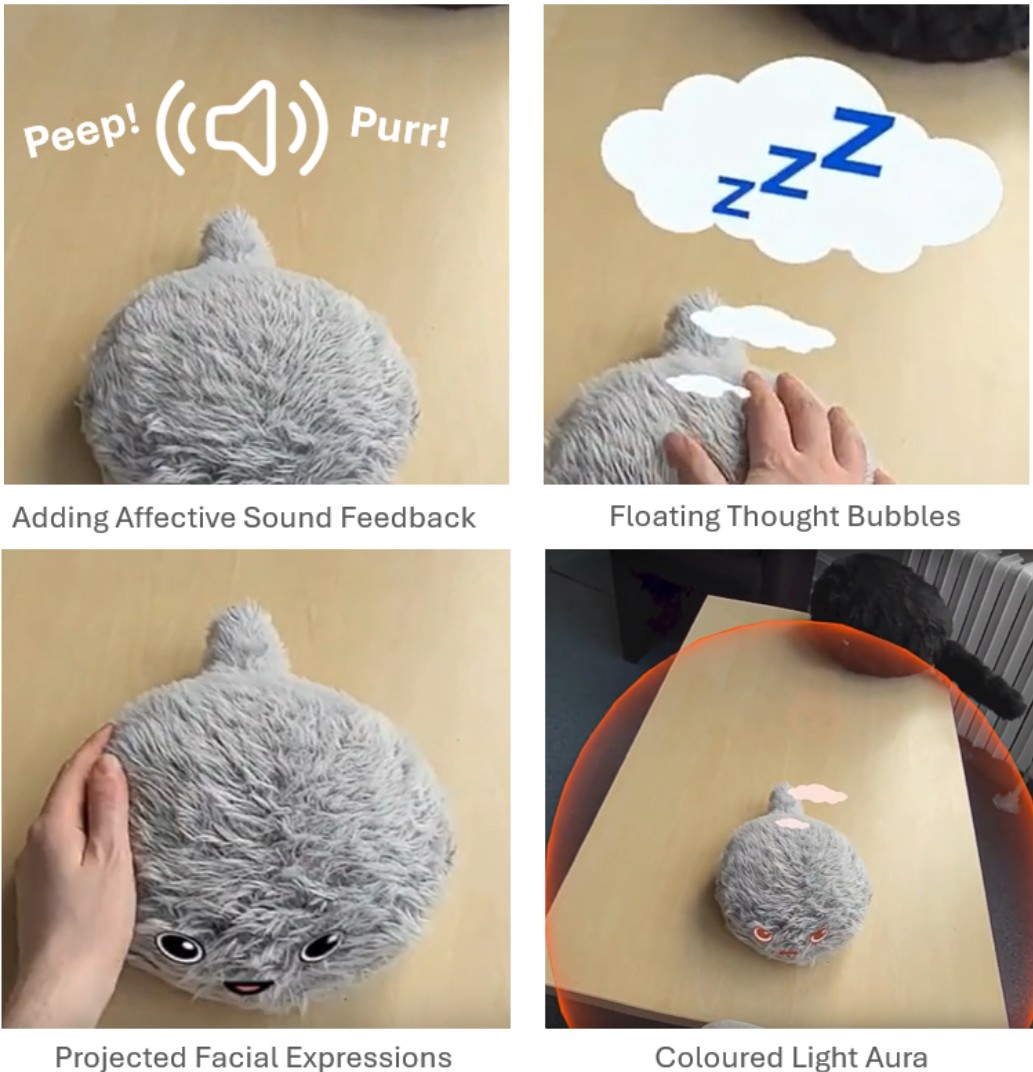}
  \caption{AZRA augments Petit Qoobo to add affective expression capabilities via four modalities: sound, a coloured aura, thought bubbles and a face.}
    \label{fig:Expressions}
\end{figure}

Applying AZRA to Petit Qoobo extends the resultant augmented robot's capacity for affective displays.
We demonstrate this by implementing four modalities drawn from prior work\cite{Loffler2018, Ghafurian2022}: facial expressions, non-verbal sound cues, coloured light and overhead thought bubbles (see Fig. \ref{fig:Expressions}).
\newline\noindent\textbf{Face:} 
Facial expressions are among the most effective modalities for enabling users to identify and empathise with emotional states in zoomorphic robots\cite{macdonald_evaluating_2024}.
We used AZRA to project simple 2D facial expressions onto the front of the robot, which was inspired by prior work using eye and mouth Action Units to differentiate emotions \cite{bretin_role_2025}. 
\newline\noindent\textbf{Light:}
Using coloured light to communicate affect is prominently featured in prior work ~\cite{Ghafurian2022, Loffler2018} and could enhance the physical distance at which a robot can express itself by affecting environmental ambience.
We emulated this in AR, projecting a sphere of coloured light around the robot.
\newline\noindent\textbf{Sound:}
Adding audio feedback can allow robots to express multisensory affective displays or states, as well as engage users when not in eyesight.
We added positional audio centred on the robot and implemented an initial set of five distinct audio cues drawn from Macdonald et al. \cite{macdonald_evaluating_2024}.
\newline\noindent\textbf{Emoji/Thought Bubbles:}
Emojis add informational or emotional context during computer-mediated communication and could serve a similar purpose for robots, providing emotional feedback or contextualising intentions or reactions. 
For example, a sleeping emoji could indicate the robot needs to rest or recharge, while a hand emoji shown alongside an emotional display could indicate the robot wants social touch.
To explore this design space, AZRA displays emojis within animated thought bubbles shown above the robot.

\begin{figure}[h!]
  \centering
  \includegraphics[width=0.92\linewidth]{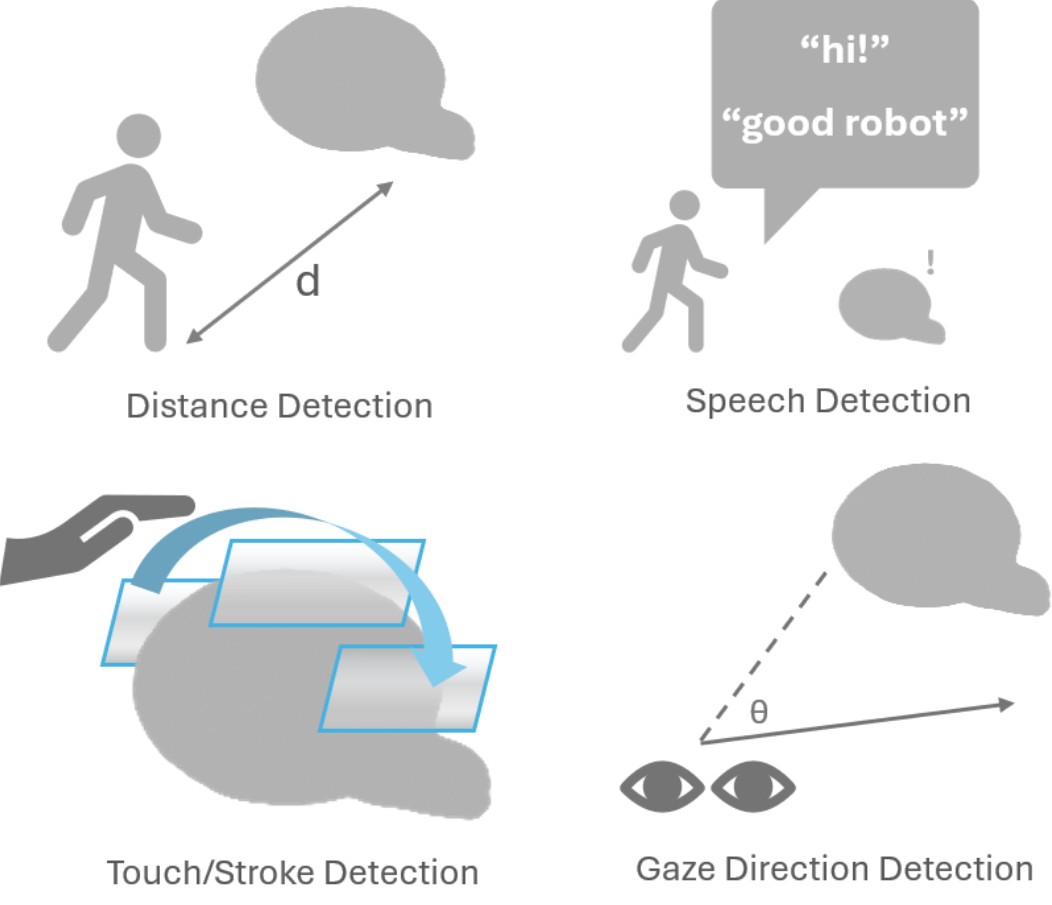}
  \caption{AZRA allows Petit Qoobo to detect four new kinds of user interaction: gaze direction, proximity, speech and directional stroking/touch.}
    \label{fig:Interactions}
\end{figure}

\begin{figure*}[h!]
  \centering
  \includegraphics[width=0.77\linewidth]{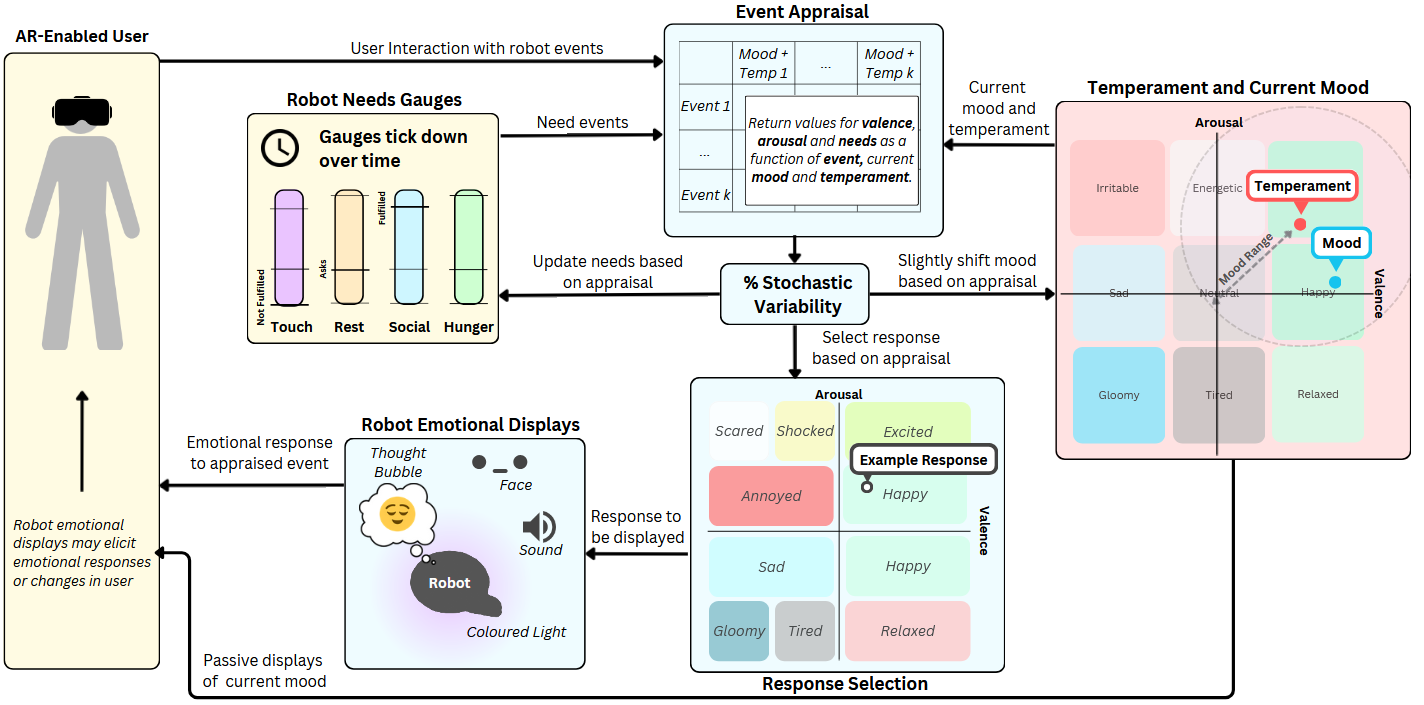}
  \caption{Architecture of the ZAMA CME, which models emotional responses to events, evolving mood/temperament and internal needs and goals.}
    \label{fig:CME}
\end{figure*}

\subsection{New Sensory Capabilities and User Interactions}

AZRA facilitates two-way interaction by implementing four new ways that the augmented Petit Qoobo can detect and react to the user: directional stroking, spoken words, gaze direction and proximity to the robot (see Fig. \ref{fig:Interactions}). 
\newline\noindent\textbf{Touch/Directional Stroking:}
Affective touch interaction is a core aspect of many human-pet relationships.
While Petit Qoobo can respond to touch in a generic sense by increasing the intensity of its tail movement, AZRA allows the augmented robot to recognise touch in different locations and track directional stroking.
This is enabled via hand-tracking and a series of colliders placed on the front, back, left, right and top of the virtual robot copy (Fig. \ref{fig:virtualcopy}).
This enables the robot to exhibit nuanced responses to touch. 
For example, the robot could respond positively when stroked front-to-back but show annoyance when stroked against the grain.
\newline\noindent\textbf{Spoken Words:}
AZRA integrates offline speech recognition\footnote{VOSK: https://alphacephei.com/vosk/}, allowing the augmented robot to recognise specific words or categories of words. 
This can enable users to communicate more complicated concepts to the robot.
For example, when a `greeting' word is recognised, such as `hello' or `hi', the robot can respond accordingly, or even contextualise its response to match using a thought bubble.
\newline\noindent\textbf{Gaze Direction:}
AZRA tracks the angle between the direction the user is facing and the robot. 
This allows the robot to react to related social cues, such as showing excitement when making eye contact with the user, or displaying feelings of loneliness when the user is looking away from them.
\newline\noindent\textbf{Proximity:}
AZRA also tracks the distance between the user and the robot, allowing the augmented robot to respond to this information. 
For example, the robot may express contentment or happiness when in prolonged close proximity to the user, feel lonely or upset when the user walks away, or express excitement and anticipation as the user gets closer.

\subsection{Modelling Robot Emotions, Temperament and Goals}
The final component of AZRA expands the affective intelligence of the augmented robot via a computational model of emotion (CME).
Most pre-existing CMEs are designed for human-like virtual agents \cite{ojha_computational_2021} rather than pet-like companions, while zoomorphic robots like Petit Qoobo or Paro primarily feature identical temperaments and limited predictable reactions to stimuli. 
To advance beyond short-term emotional interactions \cite{cavallo_emotion_2018}, we developed the Zoomorphic Robot Affect and Agency Mind Architecture (ZAMA), a CME integrated into AZRA and adapted from the pre-existing FAtiMA model \cite{dias_fatima_2014}. 
This model (see Fig. \ref{fig:CME}) governs (1) emotional expressions in response to events such as user interactions, (2) long-term temperament and mood, as well as (3) the robot's needs and subsequent actions.
Here we overview the architecture and features of this initial iteration of ZAMA, while exact figures and functionality are subject to adjustment following future evaluation.
\par
We utilised the Circumplex Model of Affect as the underlying model of emotion\cite{Russell1980}, plotting affective states on the intersection of valence and arousal axes, similarly to MOFLIN. 
The core functionality of ZAMA is similar to FAtiMA. 
Events are perceived via the interactions tracked by the HMD, such as voice or touch. 
The Event Appraisal component evaluates events as a function of the event, current mood and temperament \cite{ojha_computational_2021}, producing valence and arousal values. 
A Response Selection component then plots these values on the valence/arousal axes of the circumplex model to choose a short-term emotional response. 
This response is then sent to the Robot Emotional Displays component to determine which affective expression modalities are employed and how. For example, sound or light may be used to add emphasis for higher arousal emotional displays.
\par
Beyond this, however, several additions have been made to ZAMA to make the augmented robot more capable of long-term emotional change and dynamic independent actions.
\newline\noindent\textbf{Temperament and Mood:}
ZAMA models the robot's long-term temperament as a combination of valence and arousal values to assign the robot one of nine archetypes, such as `Gloomy', `Relaxed' or `Irritable'.
The robot's daily mood valence and arousal values are then plotted randomly within a fixed range of its current temperament (see Fig. \ref{fig:CME}).
When not expressing specific emotional responses, the robot passively displays its current mood via its AR facial expression.
\newline\noindent\textbf{Robot Needs:}
ZAMA models and tracks four internal `needs' drawn from literature on pet needs  \cite{griffin_adaptation_2023}, Touch, Rest, Socialising and Hunger, represented by gradually decreasing numerical meters.
These allow the augmented robot to simulate independence and spontaneity by initiating event appraisal and affective interaction based on its own motivations.
For example, when a lower threshold is reached in the Touch meter, the robot will appraise this event internally and show an affective display to prompt user touch, clarifying this motivation with a thought bubble that shows a hand.
\newline\noindent\textbf{Stochastic Variability:}
Stochastic variability is introduced by adding noise to valence and arousal values that result from Event Appraisal, in order to mimic the natural variations in emotional responses \cite{sanchez_affective_2021} to improve the illusion of life \cite{Kersjes2016}.
\newline\noindent\textbf{Feedback and Evolution:}
Finally, ZAMA features a feedback mechanism that allows daily changes in mood and slow changes in temperament over time, based on emotional responses to appraised events. 
For example, a robot's mood will experience a small valence increase in response to user interactions appraised with positive valence, such as stroking. 
These daily changes in mood in turn impact long-term temperament, allowing the augmented robot to slowly adapt based on user habits. 
Additionally, events that are relevant to the robot's needs also feed into those needs. 
For example, the value of the `Social Need' meter increases when the user touches the robot or is in close proximity to it.


\section{Next Steps}
Following the development of AZRA, we plan to utilise it in a series of mixed-methods iterative participatory prototyping studies.
This aims to explore how zoomorphic robots can better serve as domestic pet alternatives.
First, observing how participants interact with, perceive and seek to change AZRA in exploratory user interactions will be invaluable in refining the framework for real-world use.
In addition to allowing us to identify and address unintended or ineffective emergent behaviours, these sessions will provide formative feedback on whether the framework facilitates interaction that is understandable, emotionally engaging and evokes pet experiences.
\par
Second, we will leverage the increased scope and rapid prototyping afforded by AZRA to explore the design space of zoomorphic robotic interaction in a wider, faster and more participatory manner than previously possible. 
This may involve implementing new interactions and displays within the framework, extending the framework to explore further options, or using the flexibility of AR prototyping to explore individual personalisation and accessibility needs.
\par
Beyond this scope, AR research has used project conversational agents onto different everyday objects \cite{iwai_bringing_2025, WangApple2025}. Future work could explore the implications of this approach using specifically pet-like interaction, augmenting both related objects (e.g., plush toys) and unrelated objects.


\bibliographystyle{IEEEtran}
\bibliography{bibliography}

\begin{thebibliography}{10}
\providecommand{\url}[1]{#1}
\csname url@samestyle\endcsname
\providecommand{\newblock}{\relax}
\providecommand{\bibinfo}[2]{#2}
\providecommand{\BIBentrySTDinterwordspacing}{\spaceskip=0pt\relax}
\providecommand{\BIBentryALTinterwordstretchfactor}{4}
\providecommand{\BIBentryALTinterwordspacing}{\spaceskip=\fontdimen2\font plus
\BIBentryALTinterwordstretchfactor\fontdimen3\font minus \fontdimen4\font\relax}
\providecommand{\BIBforeignlanguage}[2]{{%
\expandafter\ifx\csname l@#1\endcsname\relax
\typeout{** WARNING: IEEEtran.bst: No hyphenation pattern has been}%
\typeout{** loaded for the language `#1'. Using the pattern for}%
\typeout{** the default language instead.}%
\else
\language=\csname l@#1\endcsname
\fi
#2}}
\providecommand{\BIBdecl}{\relax}
\BIBdecl

\bibitem{borgstedt_soothing_2024}
J.~Borgstedt, S.~Macdonald, K.~Marky, F.~Pollick, and S.~Brewster, ``Soothing {Sensations}: {Enhancing} {Interactions} with a {Socially} {Assistive} {Robot} through {Vibrotactile} {Heartbeats}.''\hskip 1em plus 0.5em minus 0.4em\relax IEEE, 2024.

\bibitem{Wada2007}
K.~Wada and T.~Shibata, ``Living with seal robots - {Its} sociopsychological and physiological influences on the elderly at a care house,'' \emph{IEEE Transactions on Robotics}, vol.~23, no.~5, pp. 972--980, 2007.

\bibitem{Young2011a}
J.~E. Young, J.~Sung, A.~Voida, E.~Sharlin, T.~Igarashi, H.~I. Christensen, and R.~E. Grinter, ``Evaluating human-robot interaction: Focusing on the holistic interaction experience,'' \emph{International Journal of Social Robotics}, vol.~3, pp. 53--67, 2011.

\bibitem{Katsuno2022}
H.~Katsuno and D.~White, ``Haptic {Creatures}: {Tactile} {Affect} and {Human}– {Robot} {Intimacy} in {Japan},'' in \emph{Consumer {Culture} {Theory} in {Asia}}, 2022, pp. 242--262.

\bibitem{Sefidgar2016}
Y.~S. Sefidgar, K.~E. MacLean, S.~Yohanan, H.~F.~H. Van Der~Loos, E.~A. Croft, and E.~J. Garland, ``Design and {Evaluation} of a {Touch}-{Centered} {Calming} {Interaction} with a {Social} {Robot},'' \emph{IEEE Transactions on Affective Computing}, vol.~7, no.~2, 2016.

\bibitem{Hudson2020}
J.~Hudson, R.~Ungar, L.~Albright, R.~Tkatch, J.~Schaeffer, and E.~R. Wicker, ``Robotic {Pet} {Use} {Among} {Community}-{Dwelling} {Older} {Adults},'' \emph{The journals of gerontology. Series B, Psychological sciences and social sciences}, vol.~75, no.~9, pp. 2018--2028, 2020.

\bibitem{Jeong2018}
S.~Jeong, C.~Breazeal, D.~Logan, and P.~Weinstock, ``Huggable: {The} impact of embodiment on promoting socio-emotional interactions for young pediatric inpatients,'' \emph{Conference on Human Factors in Computing Systems - Proceedings}, vol. 2018-April, pp. 1--13, 2018.

\bibitem{Tanevska2020}
A.~Tanevska, F.~Rea, G.~Sandini, L.~Cañamero, and A.~Sciutti, ``A {Socially} {Adaptable} {Framework} for {Human}-{Robot} {Interaction},'' \emph{Frontiers in Robotics and AI}, vol.~7, Oct. 2020.

\bibitem{cavallo_emotion_2018}
F.~Cavallo, F.~Semeraro, L.~Fiorini, G.~Magyar, P.~Sinčák, and P.~Dario, ``Emotion {Modelling} for {Social} {Robotics} {Applications}: {A} {Review},'' \emph{Journal of Bionic Engineering}, vol.~15, no.~2, pp. 185--203, 2018.

\bibitem{Loffler2018}
D.~Löffler, N.~Schmidt, and R.~Tscharn, ``Multimodal {Expression} of {Artificial} {Emotion} in {Social} {Robots} {Using} {Color}, {Motion} and {Sound},'' in \emph{{ACM}/{IEEE} {International} {Conference} on {Human}-{Robot} {Interaction}}.\hskip 1em plus 0.5em minus 0.4em\relax IEEE Computer Society, Feb. 2018, pp. 334--343, iSSN: 21672148.

\bibitem{macdonald_evaluating_2024}
\BIBentryALTinterwordspacing
S.~Macdonald, R.~Bretin, and S.~ElSayed, ``Evaluating {Transferable} {Emotion} {Expressions} for {Zoomorphic} {Social} {Robots} using {VR} {Prototyping},'' in \emph{2024 {IEEE} {International} {Symposium} on {Mixed} and {Augmented} {Reality} ({ISMAR})}, Oct. 2024, pp. 1087--1096, iSSN: 2473-0726. [Online]. Available: \url{https://ieeexplore.ieee.org/abstract/document/10765384}
\BIBentrySTDinterwordspacing

\bibitem{Suzuki2022}
R.~Suzuki, A.~Karim, T.~Xia, H.~Hedayati, and N.~Marquardt, ``Augmented {Reality} and {Robotics}: {A} {Survey} and {Taxonomy} for {AR}-enhanced {Human}-{Robot} {Interaction} and {Robotic} {Interfaces},'' in \emph{Conference on {Human} {Factors} in {Computing} {Systems} - {Proceedings}}.\hskip 1em plus 0.5em minus 0.4em\relax Association for Computing Machinery, Apr. 2022, arXiv: 2203.03254.

\bibitem{dias_fatima_2014}
\BIBentryALTinterwordspacing
J.~Dias, S.~Mascarenhas, and A.~Paiva, ``\BIBforeignlanguage{en}{{FAtiMA} {Modular}: {Towards} an {Agent} {Architecture} with a {Generic} {Appraisal} {Framework}},'' in \emph{\BIBforeignlanguage{en}{Emotion {Modeling}: {Towards} {Pragmatic} {Computational} {Models} of {Affective} {Processes}}}, T.~Bosse, J.~Broekens, J.~Dias, and J.~van~der Zwaan, Eds.\hskip 1em plus 0.5em minus 0.4em\relax Cham: Springer International Publishing, 2014, pp. 44--56. [Online]. Available: \url{https://doi.org/10.1007/978-3-319-12973-0_3}
\BIBentrySTDinterwordspacing

\bibitem{Ghafurian2022}
\BIBentryALTinterwordspacing
M.~Ghafurian, G.~Lakatos, and K.~Dautenhahn, ``The {Zoomorphic} {Miro} {Robot}’s {Affective} {Expression} {Design} and {Perceived} {Appearance},'' \emph{International Journal of Social Robotics}, vol.~14, no.~4, pp. 945--962, 2022, publisher: Springer Netherlands. [Online]. Available: \url{https://doi.org/10.1007/s12369-021-00832-3}
\BIBentrySTDinterwordspacing

\bibitem{bretin_role_2025}
\BIBentryALTinterwordspacing
R.~Bretin, M.~Khamis, E.~Cross, and M.~Obaid, ``The {Role} of {Drone}’s {Digital} {Facial} {Emotions} and {Gaze} in {Shaping} {Individuals}’ {Social} {Proxemics} and {Interpretation},'' \emph{J. Hum.-Robot Interact.}, Feb. 2025, just Accepted. [Online]. Available: \url{https://dl.acm.org/doi/10.1145/3714477}
\BIBentrySTDinterwordspacing

\bibitem{ojha_computational_2021}
\BIBentryALTinterwordspacing
S.~Ojha, J.~Vitale, and M.-A. Williams, ``\BIBforeignlanguage{en}{Computational {Emotion} {Models}: {A} {Thematic} {Review}},'' \emph{\BIBforeignlanguage{en}{International Journal of Social Robotics}}, vol.~13, no.~6, pp. 1253--1279, Sep. 2021. [Online]. Available: \url{https://doi.org/10.1007/s12369-020-00713-1}
\BIBentrySTDinterwordspacing

\bibitem{Russell1980}
J.~A. Russell, ``A circumplex model of affect,'' \emph{Journal of Personality and Social Psychology}, vol.~39, no.~6, pp. 1161--1178, 1980.

\bibitem{griffin_adaptation_2023}
\BIBentryALTinterwordspacing
K.~E. Griffin, S.~S. Arndt, and C.~M. Vinke, ``\BIBforeignlanguage{en}{The {Adaptation} of {Maslow}’s {Hierarchy} of {Needs} to the {Hierarchy} of {Dogs}’ {Needs} {Using} a {Consensus} {Building} {Approach}},'' \emph{\BIBforeignlanguage{en}{Animals}}, vol.~13, no.~16, 2023. [Online]. Available: \url{https://www.mdpi.com/2076-2615/13/16/2620}
\BIBentrySTDinterwordspacing

\bibitem{sanchez_affective_2021}
\BIBentryALTinterwordspacing
E.~Sanchez, M.~K. Tellamekala, M.~Valstar, and G.~Tzimiropoulos, ``\BIBforeignlanguage{English}{Affective {Processes}: stochastic modelling of temporal context for emotion and facial expression recognition}.''\hskip 1em plus 0.5em minus 0.4em\relax IEEE Computer Society, Jun. 2021, pp. 9070--9080. [Online]. Available: \url{https://www.computer.org/csdl/proceedings-article/cvpr/2021/450900j070/1yeL2PXBKtq}
\BIBentrySTDinterwordspacing

\bibitem{Kersjes2016}
H.~Kersjes and P.~Spronck, ``Modeling believable game characters,'' \emph{IEEE Conference on Computatonal Intelligence and Games}, 2016.

\bibitem{iwai_bringing_2025}
\BIBentryALTinterwordspacing
N.~Iwai and F.~Matulic, ``Bringing {Everyday} {Objects} to {Life} in {Augmented} {Reality} with {AI}-{Powered} {Talking} {Characters},'' in \emph{Proceedings of the {Extended} {Abstracts} of the {CHI} {Conference} on {Human} {Factors} in {Computing} {Systems}}, ser. {CHI} {EA} '25.\hskip 1em plus 0.5em minus 0.4em\relax New York, NY, USA: Association for Computing Machinery, Apr. 2025, pp. 1--7. [Online]. Available: \url{https://dl.acm.org/doi/10.1145/3706599.3719978}
\BIBentrySTDinterwordspacing

\bibitem{WangApple2025}
\BIBentryALTinterwordspacing
Y.~Wang, Y.~Lu, S.~Yan, and X.~Shen, ``"if my apple can talk": Exploring the use of everyday objects as personalized ai agents in mixed reality,'' in \emph{Proceedings of the Extended Abstracts of the CHI Conference on Human Factors in Computing Systems}, ser. CHI EA '25.\hskip 1em plus 0.5em minus 0.4em\relax New York, NY, USA: Association for Computing Machinery, 2025. [Online]. Available: \url{https://doi.org/10.1145/3706599.3720253}
\BIBentrySTDinterwordspacing

\end{thebibliography}

\end{document}